\documentclass[conference]{IEEEtran}
\IEEEoverridecommandlockouts
\usepackage{cite}
\usepackage{amsmath,amssymb,amsfonts}
\usepackage{algorithmic}
\usepackage{graphicx}
\usepackage{textcomp}

\usepackage[ruled,vlined,noend,linesnumbered]{algorithm2e}
\usepackage{commath}

\usepackage{multirow}
\usepackage{caption}
\usepackage{subcaption}
\usepackage{url}
\usepackage{float}
\usepackage{color}
\usepackage[numbers]{natbib}

\usepackage{xcolor}
\def\BibTeX{{\rm B\kern-.05em{\sc i\kern-.025em b}\kern-.08em
    T\kern-.1667em\lower.7ex\hbox{E}\kern-.125emX}}
\begin{document}

\title{Fine-grained Anomaly Detection in Sequential Data via Counterfactual Explanations}

\author{\IEEEauthorblockN{He Cheng\IEEEauthorrefmark{1}, Depeng Xu\IEEEauthorrefmark{2}, Shuhan Yuan\IEEEauthorrefmark{1}, Xintao Wu\IEEEauthorrefmark{3}}
\IEEEauthorblockA{\IEEEauthorrefmark{1} Utah State University, Logan, USA, \IEEEauthorrefmark{2} University of North Carolina at Charlotte, Charlotte, USA, \\ \IEEEauthorrefmark{3} University of Arkansas, Fayetteville, AR\\
Email: \{he.cheng,shuhan.yuan\}@usu.edu, depeng.xu@uncc.edu, xintaowu@uark.edu
}
}

\maketitle

\begin{abstract}
    Anomaly detection in sequential data has been studied for a long time because of its potential in various applications, such as detecting abnormal system behaviors from log data. Although many  approaches can achieve good performance on anomalous sequence detection, how to identify the anomalous entries in sequences is still challenging due to a lack of information at the entry-level. In this work, we propose a novel framework called CFDet for fine-grained anomalous entry detection. CFDet leverages the idea of interpretable machine learning. Given a sequence that is detected as anomalous, we can consider anomalous entry detection as an interpretable machine learning task because identifying anomalous entries in the sequence is to provide an interpretation to the detection result. We make use of the deep support vector data description (Deep SVDD) approach to detect anomalous sequences and propose a novel counterfactual interpretation-based approach to identify anomalous entries in the sequences. Experimental results on three datasets show that CFDet can correctly detect anomalous entries.
\end{abstract}


\section{Introduction}
Anomalous sequence detection has received a lot of attention recently because of wide applications, such as detecting anomalous log sequences or user activity sequences \cite{duDeepLogAnomalyDetection2017,liuLog2vecHeterogeneousGraph2019,zhouLogAnomalyUnsupervisedDetection2019,duLifelongAnomalyDetection2019,shuhanyuanFewshotInsiderThreat2020,wangMultiScaleOneClassRecurrent2021,guo2021logbert}. For example, log messages generated by computing systems are critical resources for debugging the abnormal patterns of systems or detecting novel attacks. Identifying the anomalous log sequences generated by computing systems in a timely manner is important to build stable systems \cite{duDeepLogAnomalyDetection2017,liuLog2vecHeterogeneousGraph2019,zhouLogAnomalyUnsupervisedDetection2019,guo2021logbert}.

However, the current approaches usually focus on anomalous sequence detection while no much work targets the fine-grained anomalous entry detection. In practice, given detected anomalous sequences, it is critical for the domain users to understand why the detection model made such predictions.  Identifying the anomalous entries in the sequence can help the domain users locate the exact issues. For example, if a user activity sequence is labeled as anomalous, it is important to know which activities in the sequences are anomalous that lead to an anomalous outcome. On the other hand, due to the limited information at the entry-level, it is not straightforward to detect anomalous entries from the sequences. Especially, for the anomalies like group anomalies (consisting of a collection of two or more anomalous points) or contextual anomalies (being abnormal because of their contexts),
the traditional point anomaly detection approaches are not suitable for identifying the anomalous entries in sequences.

In this work, we propose to leverage the idea of interpretable machine learning to achieve fine-grained anomalous entry detection. This is because identifying  anomalous entries from a detected anomalous sequence is to explain why the given sequence is labeled as anomalous. Specially, we propose a counterfactual explanation approach, called CFDet, to detect anomalous entries via searching for counterfactual samples. The idea of counterfactual explanation is to identify the ``smallest change'' to anomalous sequences that could change the prediction to normal. Here, the changes indicate anomalous entries, because by removing the anomalous entries, we can ``change'' a sequence from anomalous to normal. Then, we can achieve fine-grained anomalous entry detection. 

One challenge of leveraging interpretable machine learning techniques is that current interpretable models are usually developed for supervised learning models \cite{rudinInterpretableMachineLearning2021}. For the counterfactual explanation approach, a classifier is usually trained to distinguish the original samples from the counterfactual samples. However, in the anomaly detection scenario, due to limited anomalous samples, it is hard to train a supervised classification model, and most anomaly detection approaches are unsupervised or one-class classification models \cite{duDeepLogAnomalyDetection2017,ruffDeepOneClassClassification2018,ruffDeepSemiSupervisedAnomaly2020}. To tackle this challenge, we propose a framework only using the normal samples, where the counterfactuals are generated based on the distances to normal samples. In particular, we can divide an anomalous sequence into two parts, a subsequence with normal entries and a subsequence with anomalous entries. We consider the subsequence with normal entries as the counterfactual sample of the original anomalous sequence by imagining that the anomalous entries had not occurred.
Inspired by the deep support vector data description (Deep SVDD) \cite{ruffDeepOneClassClassification2018}, where the basic assumption is that the normal samples enclose to the center of a hypersphere, we aim at identifying the subsequence with anomalous entries having a large distance to the center while the counterfactual is close to the center. 

The main contributions of this paper are as follows. First, we propose CFDet, a novel anomalous sequence and fine-grained entry detection framework only using normal sequences for training. Considering the normal sequences are usually easy to obtain, our framework meets the requirements of real-world scenarios. Second, we develop a novel anomalous entry detection approach  based on the idea of counterfactual explanations, which considers anomalous entry detection as a task of providing interpretations to the detected anomalous sequences. Third, the experimental results on three datasets show that CFDet can detect anomalous sequences as well as fine-grained anomalous entries with high accuracy. 

\section{Related Work}
{\bf\noindent Anomaly Detection in Sequential Data.}
As sequential data become more and more ubiquitous, such as time series, video frames, or event data, sequential anomaly detection plays an important role in a wide spectrum of application scenarios \cite{ruff2021unifying,blazquez2021review,soldani2022anomaly,santhoshk2020anomaly}. Currently, due to a limited number of anomalies, many unsupervised or one-class deep learning approaches are proposed to detect anomalous sequences by identifying the differences between normal and anomalous patterns \cite{duDeepLogAnomalyDetection2017,zhouLogAnomalyUnsupervisedDetection2019,liuLog2vecHeterogeneousGraph2019,zhangRobustLogbasedAnomaly2019}. A typical idea is to make use of recurrent neural networks (RNNs) to capture the normal patterns from normal sequences. Then, an anomalous sequence can be detected with deviate patterns \cite{duDeepLogAnomalyDetection2017,zhouLogAnomalyUnsupervisedDetection2019}. For example, DeepLog \cite{duDeepLogAnomalyDetection2017} is trained to predict the log entry by an RNN model based on a large number of normal sequences so that RNN is able to capture the normal patterns of sequences. The anomalous sequence can then be detected when the RNN cannot correctly predict the log entries, meaning the sequence does not follow the normal patterns.
However, the majority of approaches proposed so far only focus on detecting the anomalous sequences and cannot point out fine-grained subsequences or entries in the sequences that actually lead to the anomalous outcomes. 

{\bf\noindent Interpretable Anomaly Detection.}
Interpretability in machine learning is crucial for high-stakes decisions and troubleshooting \cite{rudinInterpretableMachineLearning2021}. Interpretable machine learning techniques can be categorized into two types, intrinsic interpretability and post-hoc interpretability \cite{chenThisLooksThat2019}. Intrinsic interpretability indicates self-explanatory models that achieve interpretability directly based on their structures, while post-hoc interpretability means the interpretability is achieved by applying another model to provide explanations. There are two typical approaches to achieve the post-hoc interpretability, perturbation-based and gradient-based approaches.
The perturbation-based approaches find the important features based on their impact on the decision outcome by perturbation functions, such as LIME and SHAP \cite{ribeiroWhyShouldTrust2016,lundbergUnifiedApproachInterpreting2017}, while the gradient-based approaches identify the important features based on the gradient magnitudes, such as Grad-CAM and Integrated Gradients \cite{selvarajuGradCAMVisualExplanations2017,sundararajanAxiomaticAttributionDeep2017}. Many sequential anomaly detection models are deployed on safety-critical systems. Hence, once anomalous behaviors are detected, understanding them is imperative for the domain users to locate the problems. 

Only a few studies target interpretable anomaly detection \cite{brownRecurrentNeuralNetwork2018,liznerskiExplainableDeepOneClass2021}. To achieve intrinsic interpretation, the explainable deep one-class classification model \cite{liznerskiExplainableDeepOneClass2021} provides intrinsic interpretability for anomaly detection on image data but cannot identify the discrete anomalous entries in sequences. Meanwhile, the attention mechanism, which also provide intrinsic interpretation based on the attention weights, is also adopted    fdetecting anomalous events from sequential data \cite{brownRecurrentNeuralNetwork2018}. However, the attention scores derived in the proposed approach indicate the contributions to predicting the next event in the sequence and are not strictly related to the anomalous outcome. Some studies also achieve the post-hoc interpretation based on the perturbation-based or gradient-based interpretation approaches. Research in  \cite{antwargExplainingAnomaliesDetected2020} develops interpretable autoencoder models to identify features leading to high reconstruction errors using Shapley values. Similarly, research in  \cite{nguyenGEEGradientbasedExplainable2019} adopts variational autoencoder as the anomaly detection model and identifies important features based on the gradient values. OmniAnomaly achieves the interpretable anomaly detection for multivariate time series data by a neural network combining GRU and VAE, where the interpretation is achieved based on the reconstruction probability of each dimension of input data \cite{suRobustAnomalyDetection2019}. In this work, we target on detecting anomalies in discrete sequence data, it is hard to leverage the gradient-based approaches to achieve interpretation. We leverage the idea of counterfactual interpretation to achieve fine-grained anomaly detection, which provides human-understandable post-hoc interpretations to anomalous sequence detection.

\begin{figure*}[ht!]
    \centering
    \includegraphics[width = 0.8\textwidth]{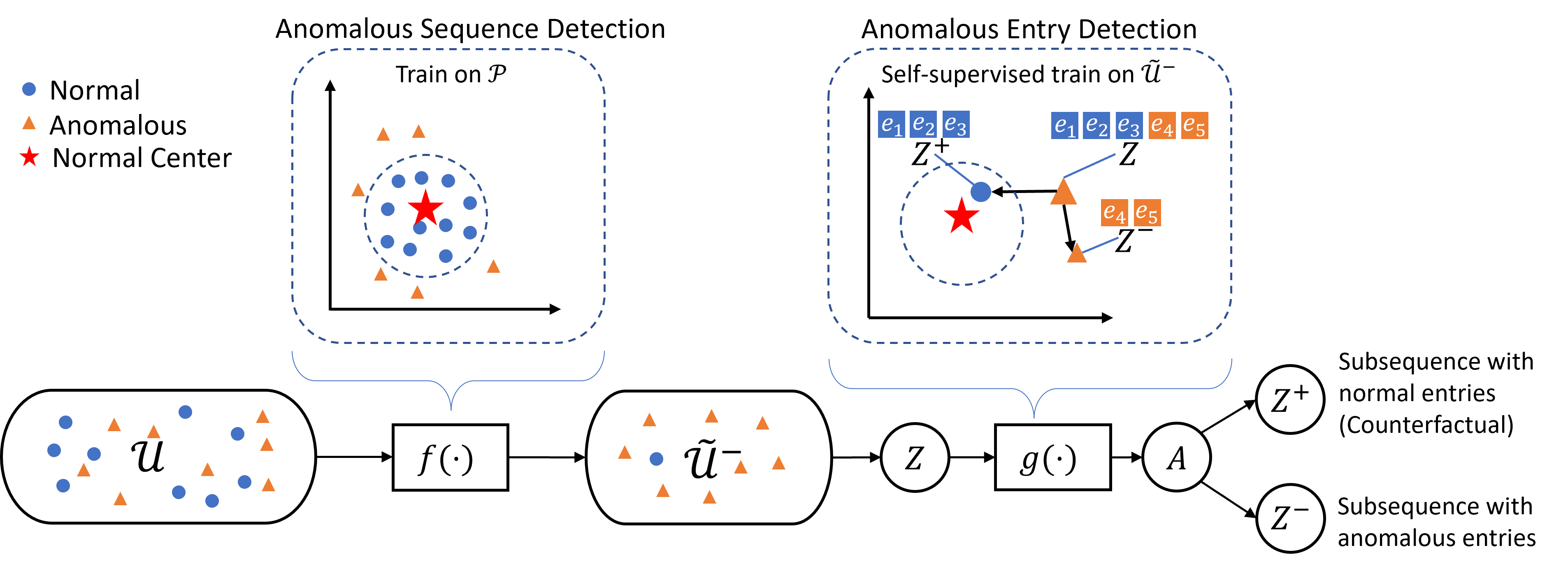}
    \caption{Illustration of CFDet for anomalous sequence and entry detection.}
    \label{fig:framework}
\end{figure*}

\section{Framework}

\subsection{Overview}

We denote a sequence with length $L$ as $S = \{e_{l}\}_{l=1}^L$ where $e_{l}$ indicates the $l$-th entry.  We use $e^+$ and $e^-$ to denote normal and anomalous entries respectively. In this work, we assume that no labeled anomalous sequences/entries are available as training signals.  Formally, given a set of normal sequences $\mathcal{P}=\{S^+_{n}\}_{n=1}^{N}$ and another set of unlabeled sequences $\mathcal{U}=\{S_m\}_{m=1}^M$ with a mixture of normal and anomalous sequences, i.e., $\mathcal{U}=\mathcal{U}^+ \cup \mathcal{U}^-$, we aim at detecting the anomalous sequences  in $\mathcal{U}$ as well as their corresponding anomalous entries. 

We propose a two-phase framework called CFDet, as shown in Figure \ref{fig:framework}. 
We first adopt the Deep SVDD approach to derive an anomalous sequence detector $f(\cdot)$ based on the normal sequence set $\mathcal{P}$. Specifically, Deep SVDD is to minimize the volume of a data-enclosing hypersphere in a latent space with a center point $\mathbf{c}$ based on $\mathcal{P}$. Therefore, the anomalies can be detected with a large distance to the center. Then, we deploy the detector $f(\cdot)$ to classify the sequences in the unlabeled set $\mathcal{U}$ into a subset of anomalous sequences  $\tilde{\mathcal{U}}^{-}$ and a subset of normal sequences $\tilde{\mathcal{U}}^{+}$. For each detected anomalous sequence set $Z \in \tilde{\mathcal{U}}^{-}$, we further identify the fine-grained anomalous entries. We propose a novel self-supervised learning approach based on the idea of counterfactual explanation to train an anomalous entry detector $g(\cdot)$. Then, we are able to identify the anomalous entries from the anomalous sequences. 

{\bf\noindent The Key Idea of Anomalous Entry Detection.}
Given a detected anomalous sequence $Z$, we denote the subsequence consisting of only anomalous entries in $Z$ as $Z^-$. Formally we have
\begin{equation}
\label{eq:z-}
    Z^- = A \odot Z,
\end{equation}
where $A=\{a_l\}_{l=1}^L$ is an indicator sequence with each $a_l \in \{0,1\}$ being a binary indicator, and $\odot$ indicates the element-wise product. If the entry is an anomalous one $e^-_l$, the corresponding indicator $a_l=1$; otherwise, $a_l=0$. Our goal is to learn an anomalous entry detector $g(\cdot): Z \rightarrow A$ that can properly generate the indicator sequence $A$.

To detect the fine-grained anomalous entries from sequences, we leverage the idea of counterfactual explanations. Counterfactual explanations describe a situation in the form: ``If X had not occurred, Y would not have occurred'' \cite{molnarInterpretableMachineLearning2019}. In the anomaly detection scenario, we can rephrase the above statement as: if there had been no anomalous entries in a sequence, the sequence would not be anomalous. 
For the anomalous sequence $Z$, we also denote its subsequence with normal entries as $Z^+$, i.e., $Z^+=(1-A) \odot Z$. Following the notion of counterfactual explanation, we aim at identifying and removing all the anomalous entries from the sequence $Z$. Once the anomalous entries are removed from the sequence, the corresponding counterfactual sequence should be normal and is actually the normal subsequence $Z^+$. 

{\bf\noindent Properties of Counterfactual Sequences.}
Because given a sequence $Z$, $Z^+$ is the complement of $Z^-$, generating a good counterfactual sequence also means achieving anomalous entry detection. An ideal counterfactual sequence of an anomalous sequence should satisfy the following properties. 

$\bullet$ \textbf{Normality:} The counterfactual sequence $Z^+$ should be normal. As illustrated in \cite{ruffDeepSemiSupervisedAnomaly2020}, by assuming that the latent representations derived by $f(\cdot)$ follow an isotropic Gaussian distribution, Deep SVDD is to minimize the upper bound on the entropy of the Gaussian. The hidden representations of counterfactual sequences should have minimal entropy:
\begin{equation}
\label{eq:normality}
    \mathcal{H}(f(Z^+)) \leq \tau,
\end{equation}
where $\mathcal{H}(\cdot)$ indicates the entropy and $\tau$ is a constant. 

$\bullet$ \textbf{Comprehensiveness:} After removing $Z^+$ from $Z$, $Z^-$ should be abnormal:
\begin{equation}
\label{eq:comprehensiveness}
    \mathcal{H}(f(Z^-)) \geq \mathcal{H}(f(Z^+))+t,
\end{equation}
where $t$ is a constant margin. It requires $Z^-$ as anomalous entries in a sequence $Z$ should have a higher entropy with a margin compared with the counterfactual sequence. Therefore, we can ensure that $Z^+$ includes all the normal entries while $Z^-$ has all the anomalous entries.

$\bullet$ \textbf{Conciseness:} The counterfactual sample means the minimum changes on the original sequence, so the anomalous entries $Z^-$ should be consecutive and sparse, 
\begin{equation}
\label{eq:conciseness}
    \sum_l |a_l - a_{l-1}| \leq c  \quad \sum_l a_l \leq s,
\end{equation}
where  $c$ and $s$ are both constants.

Both $Z^+$ and $Z^-$ can be derived based on the indicator sequence $A$ that is generated by the anomalous entry detector $g(\cdot)$. Therefore, given an anomalous sequence $Z$, the objective of  training $g(\cdot)$ is to meet the above three properties of counterfactual sequences. 

\subsection{Anomalous Sequence Detection}
Given a set of normal sequences $\mathcal{P}=\{S^+_{n}\}_{n=1}^{N}$, we derive an anomaly detection model based on the idea of Deep SVDD \cite{ruffDeepOneClassClassification2018}. First, given a normal sequence $S_n^+$, we adopt a long short-term memory (LSTM) neural network to encode the sequence into an embedding space and derive its representation as $\mathbf{r}_n^+=f(S_n^+)$, where $f(\cdot)$ indicates the LSTM model and $\mathbf{r}_n^+$ is the last hidden state of LSTM. In order to train the LSTM model, Deep SVDD  minimizes the volume of a hypersphere that encloses normal data. In other words, Deep SVDD aims at making the normal data close to the center of the hypersphere. We derive the center $\mathbf{c}$ of normal sequences in $\mathcal{P}$ by a mean operation, i.e., $\mathbf{c}=Mean(\mathbf{r}_n^+)$. Formally, the objective function of Deep SVDD is defined as:
\begin{equation}
\label{eq:deepsvdd}
    \mathcal{L}_{SVDD} = \frac{1}{N} \sum_{n=1}^{N} \norm{f(\mathbf{r}^+_n;\Theta)-\mathbf{c}}_2^2 + \lambda \norm{\Theta}_F^2,
\end{equation}
where $\Theta$ denotes the parameters in the LSTM model. Deep SVDD employs a quadratic loss for penalizing distances of normal data representations to the center $\mathbf{c}$. Therefore, it can jointly learn the LSTM model together with minimizing the volume of a data-enclosing hypersphere in the latent space. To ensure the normal sequences close to the center, the LSTM model is to extract the common factors of normal sequences. After training on the normal sequences $\mathcal{P}$, the LSTM model is able to map the normal sequences close to the center of the hypersphere and also make the anomalous sequences with deviate patterns far from the center. 

Therefore, we can  deploy the LSTM model $f(\cdot)$ to detect the anomalous sequences in the unlabeled set $\mathcal{U}$. Given a sequence $S_m \in \mathcal{U}$ and its representation $\mathbf{r}_m=f(S_m)$, we can derive the anomaly score as the distance of a sequence to the center of the hypersphere:
\begin{equation}
\label{eq:score}
    s(\mathbf{r}_m) = \norm{f(\mathbf{r}_m)-\mathbf{c}}_2^2.
\end{equation}
If the representation of the sequence $\mathbf{r}_m$ falls outside the hypersphere, i.e., $s(\mathbf{r}_m) > \epsilon$, where $\epsilon$ indicates the threshold, we will label the sequence $S_m$ as anomalous. Then, we can obtain a set of sequence $\tilde{\mathcal{U}}^{-}$ with detected anomalous sequences from $\mathcal{U}$.

\subsection{Anomalous Entry Detection}
After obtaining a set of detected anomalous sequences $\tilde{\mathcal{U}}^{-}$, we further aim at detecting the anomalous entries in the sequences. In this work, we propose to achieve anomalous entry detection via counterfactual explanations. 
In the anomaly detection scenario, the counterfactual example of an anomalous sequence is its subsequence with normal entries, so the counterfactual example and the subsequence with anomalous entries are complementary to each other. In other words, following the idea of counterfactual explanations, we can generate the counterfactual of an anomalous sequence by detecting the anomalous entries.

Given an anomalous sequence $Z \in \tilde{\mathcal{U}}^{-}$, in order to generate the counterfactual with normal entries $Z^+$ as well as the subsequence with anomalous entries $Z^-$, we propose to train the anomalous entry detector $g(\cdot)$ to generate the indicator sequence $A$, i.e., $A = g(Z)$. We use another LSTM model as a part of the implementation of $g(\cdot)$. Based on the LSTM model, we can derive the hidden state $\mathbf{h}_l$ for each entry $e_l \in Z$. Then, we apply a logistic regression model $q(\cdot)$ on $\mathbf{h}_l$ to predict the probability $p_l$ of the entry $e_l$ as anomalous: 
\begin{equation}
    \label{eq:entry}
    \mathbf{h}_l = LSTM(\mathbf{e}_l, \mathbf{h}_{l-1}) \quad p_l = q(\mathbf{h}_l), 
\end{equation}
where $\mathbf{e}_l$ denotes the representation of entry $e_l$. After rounding the probability $p_l$ to the 0-1 binary value, we get the indicator $a_l$ for the entry $e_l$:
\begin{equation}
    a_l= 
        \begin{cases}
            1, & \text{if } p_l \geq 0.5\\
            0, & \text{otherwise}.
        \end{cases}
\label{eq:a_l}
\end{equation}
Equations \ref{eq:entry} and \ref{eq:a_l} are the implementation of the anomalous entry detector $g(\cdot)$.

In order to train $g(\cdot)$ to accurately generate $A$ so that the counterfactual sequences meet the properties of normality, comprehensiveness, and conciseness, we train $g(\cdot)$ on the detected anomalous sequence set $\tilde{\mathcal{U}}^{-}$ by the following objective function:
\begin{equation}
\label{eq:obj}
    \mathcal{L} = \mathcal{L}_{n} + \alpha \mathcal{L}_{t} + \beta \mathcal{L}_{c} +\gamma \mathcal{L}_{s},
\end{equation}
where $\mathcal{L}_{n}$ is a Deep SVDD-based loss to ensure the normality of generated counterfactual sequences; $\mathcal{L}_{t}$ indicates the triplet loss that is to ensure the comprehensiveness; $\mathcal{L}_{c}$ indicates the continuity loss that is to ensure the continuity; $\mathcal{L}_{s}$ indicates the sparsity loss that is to ensure the generated counterfactual sample with the minimum change; $\alpha$, $\beta$, and $\gamma$ are hyperparameters to balance the weight of each loss term.

\textbf{Normality loss $\mathcal{L}_{n}$.} In order to meet the property of normality about the counterfactual sequence defined in Equation \ref{eq:normality}, given a detected anomalous sequence, the objective is to make the representation of counterfactual sequence $Z^+$ close to the center $\mathbf{c}$ (the center of the hypersphere with normal sequences). Therefore, we first derive the representations of counterfactual sequence $Z^+$ based on the LSTM model $f(\cdot)$ as  $\mathbf{r}_{z^+} = f(Z^+)$. Then, the loss is defined as
\begin{equation}
    \label{eq:l_n}
    \mathcal{L}_{n} = \norm{\mathbf{r}_{z^+}-\mathbf{c}}_2^2.
\end{equation}

\textbf{Triplet loss $\mathcal{L}_{t}$.} To meet comprehensiveness defined in Equation \ref{eq:comprehensiveness}, we also need to make sure $Z^+$ include all normal entries so that $Z^-$ only consists of anomalous entries. The idea is to make the representation of the subsequence with anomalous entries $Z^-$ far from the center $\mathbf{c}$ while the representation of the counterfactual sequence $Z^+$ close to the center. To this end, we also derive the representation of anomalous entries $Z^-$ by $f(\cdot)$, i.e., $\mathbf{r}_{z^-} = f(Z^-)$. Then, we consider the center $\mathbf{c}$ as an anchor, the representation of counterfactual $\mathbf{r}_{z^+}$ as a positive sample and the representation of the anomalous subsequence $\mathbf{r}_{z^-}$ as a negative sample. The triplet loss is adopted to stipulate the comprehensiveness property:

\begin{equation}
    \mathcal{L}_{t} = \max \big\{\norm{\mathbf{c}-\mathbf{r}_{z^+}}_2^2-\norm{\mathbf{c}-\mathbf{r}_{z^-}}_2^2+\lambda_t, 0 \big\},
    \label{eq:triplet}
\end{equation}
where $\lambda_t$ is a margin between positive and negative pairs. Intuitively, if the distance between $\mathbf{c}$ and $\mathbf{r}_{z^-}$ is larger than the distance between $\mathbf{c}$ and $\mathbf{r}_{z^+}$ with a margin, $\mathbf{r}_{z^-}$ should only have anomalous entries.

\textbf{Continuity loss $\mathcal{L}_{c}$.} Meanwhile, the abnormal entries in a sequence are usually  coherent. For example, if a system is under attack, the abnormal log entries are often consecutive. Hence, to ensure the generated indicator sequence $A$ with consecutive selection on the abnormal entries, inspired by \cite{yuRethinkingCooperativeRationalization2019}, we also incorporate the continuity loss:
\begin{equation}
    \label{eq:continuity}
      \mathcal{L}_c =  \max\{\sum_l |a_l - a_{l-1}|-\lambda_c,0\},
\end{equation}
where $\lambda_c$ is a hyperparameter that controls the continuity of the indicator sequence. Minimizing the continuity loss defined in Equation \ref{eq:continuity} ensures the indicator sequence $A$ with minimum number of small pieces controlled by $\lambda_c$. 

\textbf{Sparsity loss $\mathcal{L}_{s}$.} The counterfactual explanation usually expects the ``minimum'' change on the original sample. In our scenario, we expect that removing the detected anomalous entries is just enough to change the anomalous sequence to a normal one. We do not want to remove the normal entries which could lead to false positive detection. Moreover, in most scenarios, the anomalous entries should be sparse compared with the normal entries in a sequence. Hence, we also incorporate the sparsity loss in the loss function:
\begin{equation}
    \label{eq:sparsity}
    \mathcal{L}_s = \max\{\sum_l a_l - \lambda_s,0\}
\end{equation}
where $\lambda_s$ is a hyperparameter that indicates the expectation of anomalous entries in the sequence. Minimizing the sparsity loss is to make the number of detected anomalous entries close to a pre-set value.

\textbf{Training.} Because the indicator sequence $A$ is a sequence with binary values, the regular gradient descent algorithm cannot be used to optimize the anomalous entry detector $g(\cdot)$. Here, we use the policy gradient algorithm used in reinforcement learning \cite{suttonReinforcementLearningIntroduction2018} to train CFDet. 
To this end, we can consider the negative loss of the objective function in Equation \ref{eq:obj} as the reward function of a reinforcement learning model, and $g(\cdot)$ as an agent which takes an action $\{0,1\}$ on an entry $e_l$ based on the current state (i.e., the hidden representation of an entry $\mathbf{h}_l$). The anomalous entry detector is then trained to maximize the reward function. 

Algorithm \ref{algr:train} shows the pseudocode of CFDet to achieve anomalous entry detection on an unlabeled dataset by only using the normal dataset $\mathcal{P}$. CFDet consists of two training phases. First, we train anomalous sequence detector $f(\cdot)$ based on the idea of Deep SVDD on a normal data set $\mathcal{P}$ (lines \ref{algr:line:begin_f}-\ref{algr:line:end_f}). After training, we deploy $f(\cdot)$ on the unlabeled dataset $\mathcal{U}$ to compose the detected anomalous sequence dataset $\tilde{\mathcal{U}}^{-}$ (line \ref{algr:line:deploy_f}). Then, we train the anomalous entry detector $g(\cdot)$ on $\tilde{\mathcal{U}}^{-}$ (lines \ref{algr:line:begin_g}-\ref{algr:line:end_g}). Finally, we can deploy $g(\cdot)$ to detect the anomalous entries for sequences in $\tilde{\mathcal{U}}^{-}$ (line \ref{algr:line:deploy_g}). It is worth noting that for the sequences detected as normal based on $f(\cdot)$, we assume all the entries are normal.

\begin{algorithm}[h]
	\SetAlFnt{\tiny}
	\DontPrintSemicolon
	\SetKwInOut{Inputs}{Inputs}\SetKwInOut{Outputs}{Outputs}
	\Inputs{Normal Dataset $\mathcal{P}$ and Unlabeled Dataset $\mathcal{U}$}
	\Outputs{Detected anomalous sequence and entries in $\mathcal{U}$}

    \For(\tcp*[h]{epoch $k$}){$k=1 \rightarrow K_1$} 
    {  \label{algr:line:begin_f}
        Train anomalous sequence detector $f(\cdot)$ on $\mathcal{P}=\{S^+_{n}\}_{n=1}^{N}$ via the objective function Eq. \ref{eq:deepsvdd} 
    }  \label{algr:line:end_f}

    Deploy $f(\cdot)$ to detect the anomalous sequences in $\mathcal{U}$ based on Eq. \ref{eq:score} and get $\tilde{\mathcal{U}}^{-}$ \\ \label{algr:line:deploy_f}
    
    \For(\tcp*[h]{epoch $k$}){$k=1 \rightarrow K_2$} 
    { \label{algr:line:begin_g}
	    Train anomalous entry detector $g(\cdot)$ on $\tilde{\mathcal{U}}^{-}$ via the objective function Eq. \ref{eq:obj} \\ 
    } \label{algr:line:end_g}

    Deploy $g(\cdot)$ to detect anomalous entries for sequences in $\tilde{\mathcal{U}}^{-}$ \\ \label{algr:line:deploy_g}

	\Return detected anomalous sequences and entries in $\mathcal{U}$
\caption{Anomalous Sequence and Entry Detection}
\label{algr:train}
\end{algorithm}

\section{Experiments}

\subsection{Experimental Setting}

\begin{table}[]
\footnotesize
    \caption{Statistics of Three Datasets}
    \label{tb:datasets}
    \centering
    \resizebox{.48\textwidth}{!}{
    \begin{tabular}{c|c|c|c}
        \hline
        \multirow{2}{*}{Dataset} & \multirow{2}{*}{\begin{tabular}[c]{@{}c@{}}Normal Dataset \\ $\mathcal{P}$ (seq)\end{tabular}} & \multicolumn{2}{c}{Unlabeled Dataset $\mathcal{U}$}                                                                                                           \\ \cline{3-4} 
                                 &                                                                        & \begin{tabular}[c]{@{}c@{}}Normal\\ (seq)\end{tabular} & \begin{tabular}[c]{@{}c@{}}Anomalous\\ (seq/entry)\end{tabular} \\ \hline
        BGL                      & 344576                                                                & 77548                                                                 & 36470/627373                                                                  \\ 
        Thunderbird              & 280064                                                                       & 63126                                                                & 134365/408202                                                                  \\ 
        CERT                     & 1391104                                                                       & 572560                                                                 & 52033/121751                                                                  \\ \hline
        \end{tabular}
        }
    \end{table}

{\bf \noindent Datasets.}
We evaluate our model on the following three datasets, which all provide entry-level labels: 

\begin{itemize}
    \item \textbf{BlueGene/L (BGL)} \cite{oliner2007supercomputers} contains log messages collected from a BlueGeme/L supercomputer system at Lawrence Livermore National Labs. The log messages can be categorized into alert and not-alert messages. 
    \item \textbf{Thunderbird} \cite{oliner2007supercomputers} is a large-scale system log dataset that is collected from a Thunderbird supercomputer system at Sandia National Labs.
    \item \textbf{CERT Insider Threat Dataset (CERT)} \cite{glasserBridgingGapPragmatic2013} is a synthetic dataset consisting of log files that record the computer-based activities for all employees in an institution, such as logon, logoff, email, http visit. The CERT dataset contains 3995 benign employees and 5 insiders. On average, the number of activities for each employee is around 40000. We use version 4.2 of the CERT dataset.
\end{itemize}

For BGL and Thunderbird, we apply the log parser, Drain developed in \cite{heDrainOnlineLog2017}, to transfer the raw unstructured log messages to log templates and represent the log sequences as the template sequences. For CERT, we use the user activities to compose the sequences. For all three datasets, we adopt a sliding window with size 20 to split the log files into sequences and set the step size as 10.

Table \ref{tb:datasets} shows the statistics of the normal dataset $\mathcal{P}$ and unlabeled datasets $\mathcal{U}$, where the last column indicates the numbers of anomalous sequences as well as the anomalous entries in the unlabeled datasets $\mathcal{U}$. 
For BGL, Thunderbird, and CERT, the ratios of anomalous entries in anomalous sequences are 0.86, 0.15, and 0.12, respectively. It is worth noting that the ground-truth about the anomalous sequences and entries in the unlabeled dataset $\mathcal{U}$ is not available during the training phase. We also build a small validation set for each dataset to tune the hyper-parameters in CFDet as well as baselines for the anomalous sequence and entry detection, where the normal/anomalous sequences on BGL, Thunderbird, and CERT are 8617/4053, 7015/14930, 63618/5782, respectively. We deploy the models with the best performance on the validation set to detect anomalous sequences and entries on $\mathcal{U}$, and report their results throughout this section.

\begin{table*}[!ht]
\centering
\caption{Results of anomalous entry detection on the unlabeled dataset $\mathcal{U}$ (mean $\pm$ std.). All values are percentages.}
\label{tb:entry}
\begin{tabular}{l|l|ccccc}
    \hline
    Dataset& Metric                             & OCSVM                 & iForest              & Attention             & Shapley                    & CFDet \\ \hline
    \multirow{4}{*}{BGL}          & Precision   & 11.89$\pm${0.00}      & 19.47$\pm${0.00}     & 70.19$\pm${18.57}     & 98.05$\pm${0.98}           & \textbf{98.91}$\pm${0.91}   \\
                                  & Recall      & 14.33$\pm${0.00}      & 23.82$\pm${0.00}     & 76.57$\pm${17.45}     & 97.56$\pm${1.86}           & \textbf{98.56}$\pm${1.52}   \\
                                  & F-1 score   & 13.00$\pm${0.00}      & 21.43$\pm${0.00}     & 72.52$\pm${16.28}     & 97.79$\pm${0.77}           & \textbf{98.73}$\pm${0.82}   \\
                                  & AUC         & 52.93$\pm${0.00}      & 57.98$\pm${0.00}     & 81.70$\pm${10.70}     & 98.41$\pm${0.83}           & \textbf{99.07}$\pm${0.74}   \\ \hline
    \multirow{4}{*}{Thunderbird}  & Precision   & 8.26$\pm${0.00}       & 42.23$\pm${0.00}     & 33.40$\pm${28.66}     & 67.22$\pm${14.59}          & \textbf{90.99}$\pm${10.33}  \\
                                  & Recall      & 100.00$\pm${0.00}     & 99.99$\pm${0.00}     & 41.80$\pm${24.52}     & 94.17$\pm${5.33}           & \textbf{100.00}$\pm${0.00}   \\
                                  & F-1 score   & 15.27$\pm${0.00}      & 59.39$\pm${0.00}     & 31.44$\pm${16.34}     & 77.96$\pm${11.45}          & \textbf{94.97}$\pm${6.41}   \\
                                  & AUC         & 73.63$\pm${0.00}      & 96.75$\pm${0.00}     & 63.22$\pm${10.12}     & 94.17$\pm${4.13}           & \textbf{99.33}$\pm${0.97}   \\ \hline
    \multirow{4}{*}{CERT}         & Precision   & N/A                   & N/A                  & 71.35$\pm${20.02}     & 60.72$\pm${9.05}           & \textbf{99.31}$\pm${2.19}   \\
                                  & Recall      & N/A                   & N/A                  & 35.17$\pm${14.53}     & \textbf{55.41}$\pm${0.20}  & 55.31$\pm${0.10}   \\
                                  & F-1 score   & N/A                   & N/A                  & 46.33$\pm${17.52}     & 57.64$\pm${3.97}           & \textbf{71.04}$\pm${0.50}   \\
                                  & AUC         & N/A                   & N/A                  & 67.52$\pm${7.27}      & 77.52$\pm${0.04}           & \textbf{77.65}$\pm${0.04}  \\ \hline
    \end{tabular}
\end{table*}

{\bf\noindent Baselines.}
We compare our CFDet with two types of baselines that can achieve entry-level anomaly detection. The first one is interpretation-based approaches and the other is traditional anomaly detection approaches.

For interpretation-based approaches, we implement the following two baselines. 

\begin{itemize}
    \item \textbf{Attention.} The attention mechanism is often used to provide interpretations to deep learning models based on the attention weights \cite{xuShowAttendTell2015}. We improve the LSTM model $f(\cdot)$ by adding the attention mechanism as a new anomalous sequence detector and train the model based on the Deep SVDD loss. The attention weights show how much contributions the entries in sequences make for the final predictions. A threshold is set to identify the anomalous entries. 
    \item \textbf{Shapley.} The Shapley value is a classical approach that attributes the prediction of a machine-learning model on an input to its base features \cite{sundararajanManyShapleyValues2020,molnarInterpretableMachineLearning2019}. We derive the Shapley values of  entries in sequences as interpretations to the detecting results of the anomalous sequence detector. Especially, given an anomalous sequence, an entry with a high Shapley value indicates that replacing the entry with an entry in a normal sequence could significantly reduce the distance from the sequence to the normal center $\mathbf{c}$. 
\end{itemize}

We highlight that the above two approaches are not ``baselines'' in the usual case because the idea of combining a detection model with an interpretation approach to detect anomalous entries in sequences is novel and has not been studied by earlier work. 

For one-class anomaly detection approaches, we use the following two classical approaches:
\begin{itemize}
    \item \textbf{One Class Support Vector Machines (OCSVM)} is a one-class anomaly detection model trained only by known normal samples \cite{scholkopfEstimatingSupportHighDimensional2001}.
    \item \textbf{Isolation Forest (iForest)} is  a widely-used unsupervised method for outlier detection based on the ensemble of binary trees \cite{liuIsolationForest2008}.
\end{itemize}

For BGL and Thunderbird, we use words in log messages as features of each entry and build bag-of-words vectors as inputs to OCSVM and iForest. For CERT, the entries in sequences are only about the user activities, such as http visit, email, and file operation, which are hard to define proper features about these activities as inputs to OCSVM and iForest. Hence, we do not have results on the CERT dataset, shown as N/A in Table \ref{tb:entry}. This also shows the challenge of fine-grained anomaly detection in sequential data faced by traditional anomaly detection approaches, and the advantage of our approach. 

{\bf \noindent Implementation Details.}
We represent the log templates in BGL and user activities in CERT as embedding vectors with a size of 50 and in Thunderbird with a size of 500. For both BGL and CERT, we set the LSTM model used as anomalous sequence detector $f(\cdot)$ as a single-layer LSTM with a hidden size of 128, while for Thunderbird, we use a single-layer LSTM with a hidden size of 512. For all three datasets, we train $f(\cdot)$ with the Deep SVDD loss in 50 epochs and update the center $\mathbf{c}$ in the first 20 epochs. For the LSTM model used in the anomalous entry detector $g(\cdot)$, for all three datasets, we use a single layer LSTM with a hidden size as 128, which is trained in 100 epochs. The \textbf{code and datasets} are available online \footnote{\url{https://bit.ly/3dTykgj}}. 

For the attention-based baseline, we set the threshold as 0.05 because the sequence length is 20, which means if one entry makes a contribution higher than an average ratio, we will label it as anomalous. For the Shapley value-based baseline, once a sequence is detected as an anomaly, we consider the entries with positive Shapley values as anomalous entries. To ensure a fair comparison, all hyperparameters in baselines are also tuned based on the validation set.

{\bf \noindent Evaluation Metrics.} 
We adopt the precision, recall, F-1 score, and Area Under Receiver Operating Characteristic Curve (AUC) to evaluate the performance of anomalous sequence and entry detection and report the mean and standard deviation after 10 times of running. Precision, recall, and F-1 score indicate the performance focusing on the anomaly class, while AUC indicates the true positives against false positives  across various anomalous score thresholds.

\subsection{Experimental Results}

\subsubsection{The Performance of Anomalous Entry Detection}
Table \ref{tb:entry} shows the performance of detecting anomalous entries on the unlabeled dataset $\mathcal{U}$. We notice that two traditional anomaly detection models, OCSVM and iForest, cannot achieve good performance on detecting anomalous entries due to two possible reasons: the bag-of-words vectors based on the texts in log messages are not ideal features to identify anomalies, and some contextual anomalies cannot be identified solely based on the information in each entry. Two interpretation-based baselines, Attention and Shapley, achieve better performance than OCSVM and iForest, which shows the advantage of applying interpretation-based models for anomalous entry detection. Shapley can achieve very good performance on the BGL dataset. This could be because the ratio of anomalous entries in anomalous sequences is high, thus making the anomalous entries easy to be detected. Our CFDet achieves the best performance in terms of F-1 score and AUC on all three datasets. Especially, on Thunderbird, CFDet detects all the anomalous entries in the unlabeled dataset. Meanwhile, CFDet also achieves perfect precision on CERT but with low recall. We will explain the reason in the following subsection. Meanwhile, the standard deviations of our CFDet are close to zero on CERT. This is because CFDet keeps stable on detecting the rare events in the sequences.

\begin{table}[!h]
\small
    \centering
    \caption{Anomalous sequence detection (mean $\pm$ std.)}
    \label{tb:seq_results}
    \begin{tabular}{c|c|c|c}
        \hline
                                        &   BGL              &  Thunderbird       &  CERT \\ \hline
        Precision                       &   98.81$\pm${0.39}   &  94.59$\pm${3.62}    &  100.00$\pm${0.00}\\ 
        Recall                          &   98.02$\pm${1.14}   &  100.00$\pm${0.00}   &  63.77$\pm${0.00}\\ 
        F-1 score                       &   98.41$\pm${0.56}   &  97.19$\pm${1.96}    &  77.88$\pm${0.00}\\ \hline \hline
        AUC                             &   98.73$\pm${0.54}   &  93.75$\pm${4.55}    &  81.88$\pm${0.00}\\ \hline
        
        \end{tabular}
    \end{table}

\subsubsection{Subcomponent Evaluation}
Our approach consists of two components, Deep SVDD-based anomalous sequence detection and counterfactual explanation-based anomalous entry detection. We evaluate the performance of each component.

\textit{Deep SVDD-based anomalous sequence detection.}
Table \ref{tb:seq_results} shows the performance of anomalous sequence detection on the unlabeled dataset $\mathcal{U}$.
In short, on all three datasets, the anomalous sequence detector of our CFDet achieves good performance, which provides a solid foundation for the following anomalous entry detection. For BGL and Thunderbird, the anomalous sequence detector  achieves nearly perfect recall values with high precision, which means our model is able to detect all the anomalous sequences with a small number of false alerts. For CERT, all the detected anomalous sequences are truly anomalous (precision=1.0), but some anomalous sequences can not be correctly identified (recall=0.6377). This explains why the recall (0.5528) on the anomalous entry detection (shown in Table \ref{tb:entry}) is not high. 
Note that if an anomalous sequence is falsely labeled as normal, we would falsely consider all entries in the sequence are normal, thus leading to a low recall value.

\begin{table}[h!]
\centering
\caption{Counterfactual explanation-based anomalous entry detection on the detected anomalous sequences $\tilde{\mathcal{U}}^{-}$ (mean $\pm$ std.)}
\label{tb:cf}
\resizebox{.48\textwidth}{!}{
\begin{tabular}{l|l|ccccc}
    \hline
    Dataset& Metric                             & Attention             & Shapley                   & CFDet \\ \hline
    \multirow{4}{*}{BGL}          & Precision   & 70.19$\pm${18.57}     & 98.05$\pm${0.98}          & \textbf{98.91}$\pm${0.91}  \\
                                  & Recall      & 77.17$\pm${17.62}     & 98.37$\pm${1.84}          & \textbf{99.38}$\pm${1.51}  \\
                                  & F-1 score   & 72.80$\pm${16.39}     & 98.19$\pm${0.77}          & \textbf{99.13}$\pm${0.83}  \\
                                  & AUC         & 55.99$\pm${17.45}     & 93.31$\pm${2.51}          & \textbf{96.33}$\pm${2.81}  \\ \hline
    \multirow{4}{*}{Thunderbird}  & Precision   & 33.40$\pm${28.66}     & 67.22$\pm${14.59}         & \textbf{90.99}$\pm${10.33} \\
                                  & Recall      & 51.79$\pm${17.08}     & 94.17$\pm${5.33}          & \textbf{100.00}$\pm${0.00}  \\
                                  & F-1 score   & 34.61$\pm${15.20}     & 77.96$\pm${11.45}         & \textbf{94.97}$\pm${6.41}  \\
                                  & AUC         & 58.87$\pm${11.48}     & 92.82$\pm${4.88}          & \textbf{99.04}$\pm${1.38}  \\ \hline
    \multirow{4}{*}{CERT}         & Precision   & 71.35$\pm${20.02}     & 60.72$\pm${9.05}          & \textbf{99.31}$\pm${2.19}  \\
                                  & Recall      & 58.94$\pm${24.37}     & \textbf{91.60}$\pm${0.33} & 90.73$\pm${0.16}  \\
                                  & F-1 score   & 63.26$\pm${23.02}     & 72.67$\pm${6.39}          & \textbf{94.81}$\pm${0.94}  \\
                                  & AUC         & 78.20$\pm${12.25}     & 91.92$\pm${1.26}          & \textbf{95.32}$\pm${0.05} \\ \hline
    \end{tabular}
    }
\end{table}

\textit{Counterfactual explanation-based anomalous entry detection.}
We further analyze the counterfactual explanation-based anomalous entry detection. Table \ref{tb:cf} shows the anomalous entry detection on the detected anomalous sequences $\tilde{\mathcal{U}}^{-}$ whereas Table \ref{tb:entry} shows results on the whole unlabeled dataset $\mathcal{U}$. Both OCSVM and iForest directly detect anomalous entries and they do not depend on the prediction results from Deep SVDD. Hence we do not report the results of OCSVM and iForest in Table \ref{tb:cf}. 
On the set of detected anomalous sequences $\tilde{\mathcal{U}}^{-}$, CFDet achieves extremely high F-1 scores and AUC on all three datasets. It means CFDet can provide the post-hoc interpretation to the anomalous sequence detection with high fidelity. Especially, for CERT, we can conclude that once a sequence is detected as anomalous, CFDet can accurately detect its anomalous entries.  We also observe similar results on Shapley. The performance on  $\tilde{\mathcal{U}}^{-}$ is better than the unlabeled dataset $\mathcal{U}$. This shows again that a good interpretable model can identify anomalous entries by providing interpretation to the detection results. 

Based on the results shown in Tables \ref{tb:entry}, \ref{tb:seq_results}, and \ref{tb:cf}, we  conclude that CFDet is able to provide good explanations to the results of anomalous sequence detection. When the task is to identify the anomalous entries in unlabeled data, the performance of the anomalous sequence detection is crucial. 

\begin{figure}[h]
    \centering
    \begin{subfigure}[b]{0.23\textwidth}
        \centering
        \includegraphics[width=0.95\textwidth]{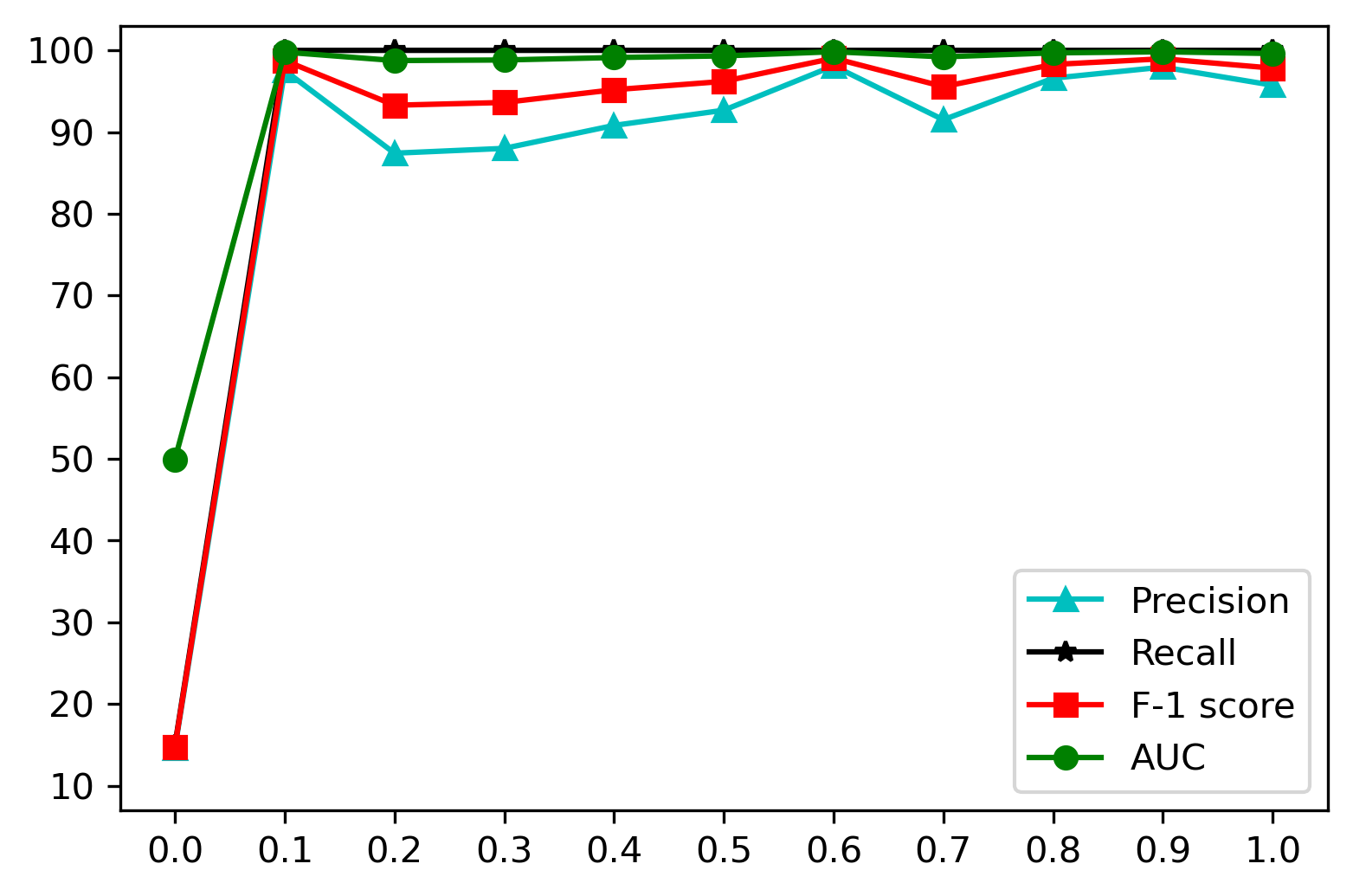}
        \caption{Triplet weight $\alpha$}
        \label{fig:alpha}
    \end{subfigure}
    \hfill
   \begin{subfigure}[b]{0.23\textwidth}
        \centering
        \includegraphics[width=0.95\textwidth]{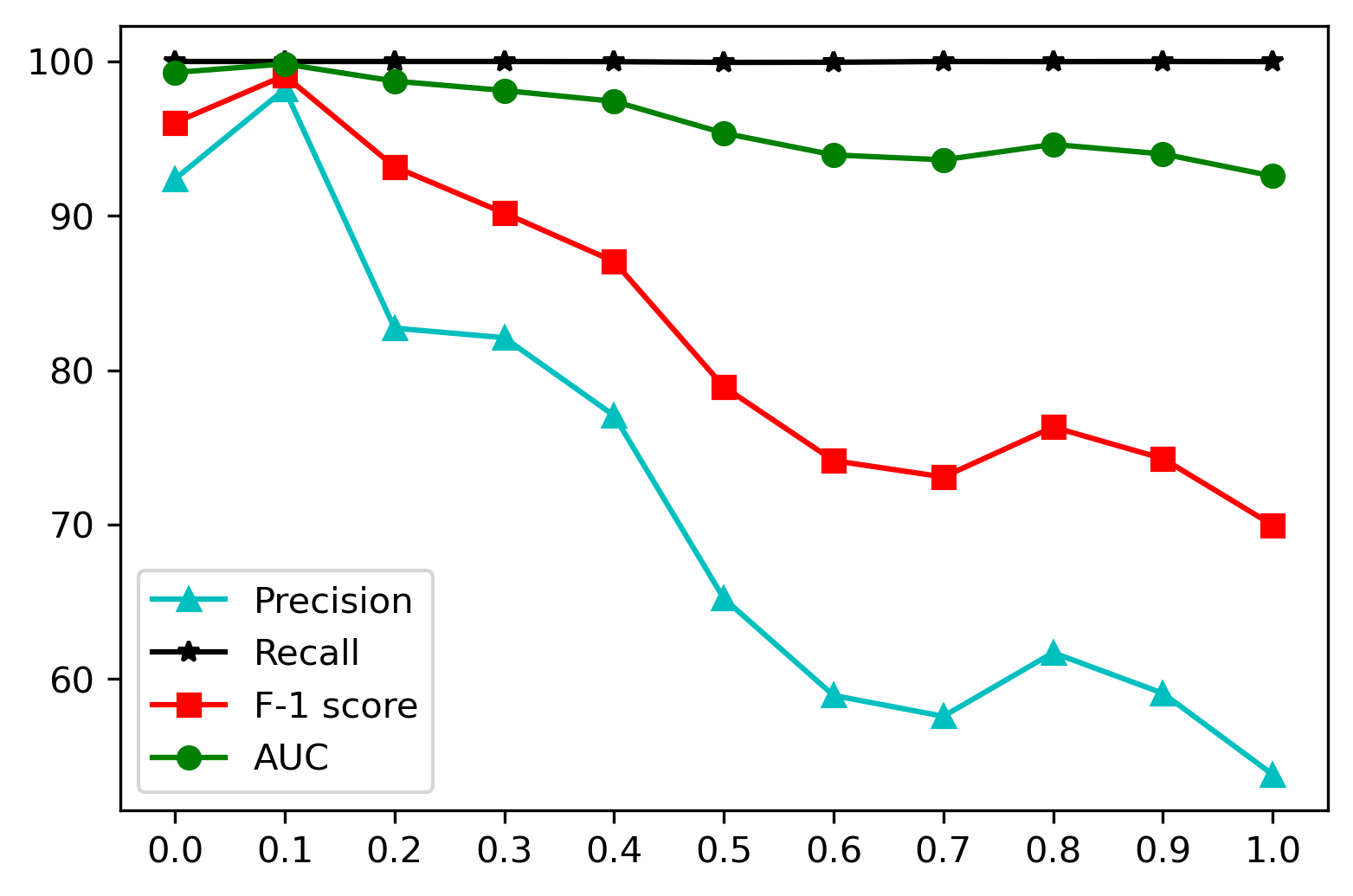}
        \caption{Continuity weight $\beta$}
        \label{fig:beta}
    \end{subfigure}
    \hfill
    \begin{subfigure}[b]{0.23\textwidth}
       \centering
       \includegraphics[width=0.95\textwidth]{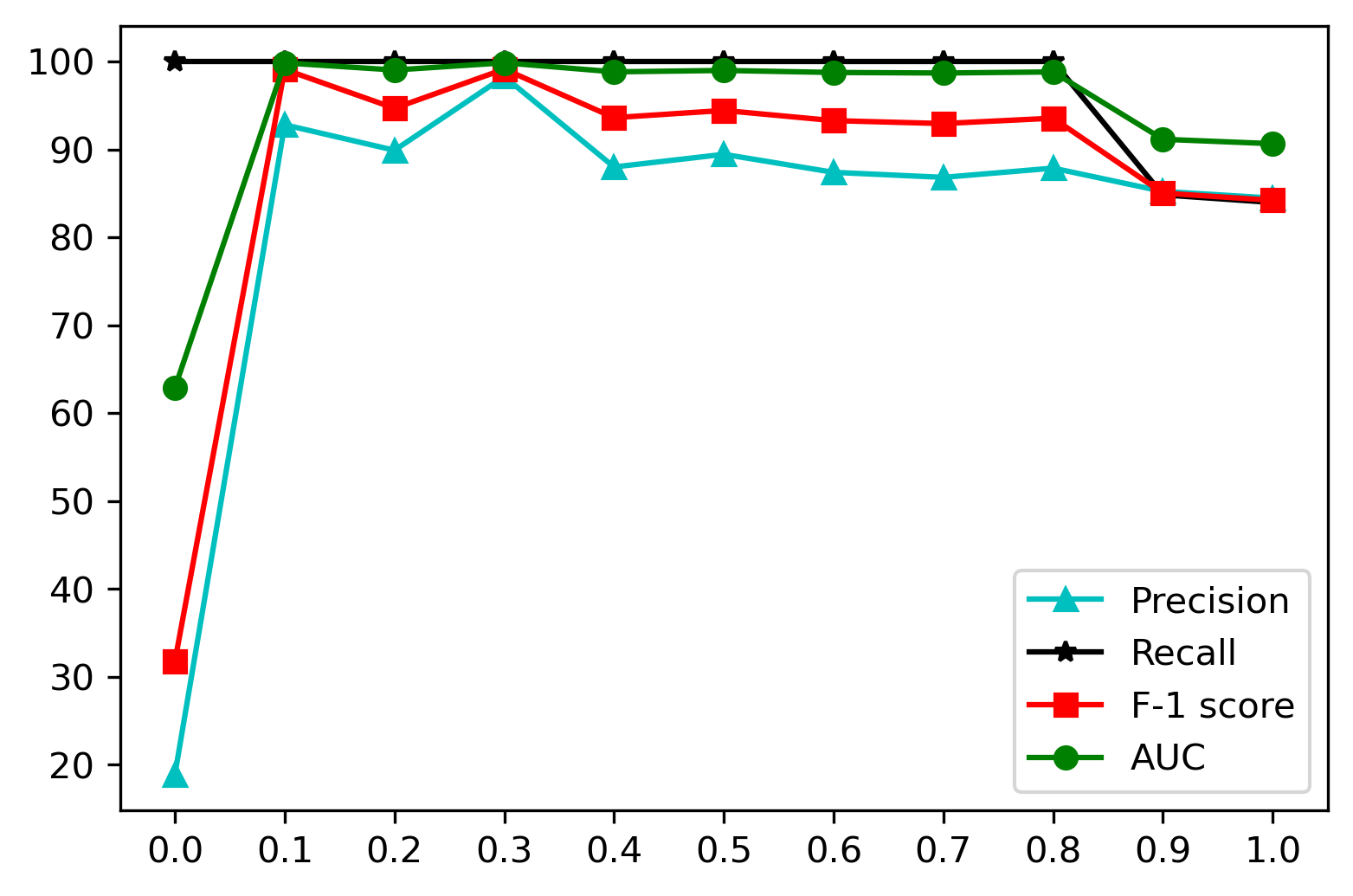}
       \caption{Sparsity weight $\gamma$}
       \label{fig:gamma}
   \end{subfigure}
   \caption{Hyperparameter sensitivity on the weight of each loss term in Equation \ref{eq:obj}}
   \label{fig:weight_obj}
\end{figure}

\subsubsection{Sensitivity Analysis}
We adopt Thunderbird as an example to analyze how triplet loss, continuity loss and sparsity loss affect anomalous entry detection. 

{\bf \noindent Sensitivity analysis on the weight of each loss term in Equation \ref{eq:obj}.}
We first analyze the effects on performance from the weights of triplet loss $\alpha$, the continuity loss $\beta$, and sparsity loss $\gamma$ defined in Equation \ref{eq:obj}.
First, Figure \ref{fig:weight_obj} shows that triplet loss, continuity loss, and sparsity loss are critical for anomalous entry detection. If we set the weight of one loss to 0 ($\alpha=0$, $\beta=0$ or $\gamma=0$), i.e., removing one loss term from the objective function, the F-1 score and AUC are low. Especially, for the triplet weight, when $\alpha=0$, the AUC value is just around 50\%, and the F-1 score is around 10\% (shown in Figure \ref{fig:alpha}), meaning the critical of triplet loss for accurately detecting anomalous entries. We can have a similar observation for the sparsity loss shown in Figure \ref{fig:gamma}, i.e., no sparsity loss ($\gamma=0$) leading to low F-1 and AUC. Meanwhile, we can also notice that both the triplet weight $\alpha$ and sparsity weight $\gamma$ do not have significant impacts on the performance as long as the values are not zero. On the other hand, the performance is sensitive to the continuity weight $\beta$. As shown in Figure \ref{fig:beta}, when the $\beta$ value keeps increasing, the performance becomes worse. 

\begin{figure}
\centering
    \begin{subfigure}[b]{0.23\textwidth}
       \centering
       \includegraphics[width=0.95\textwidth]{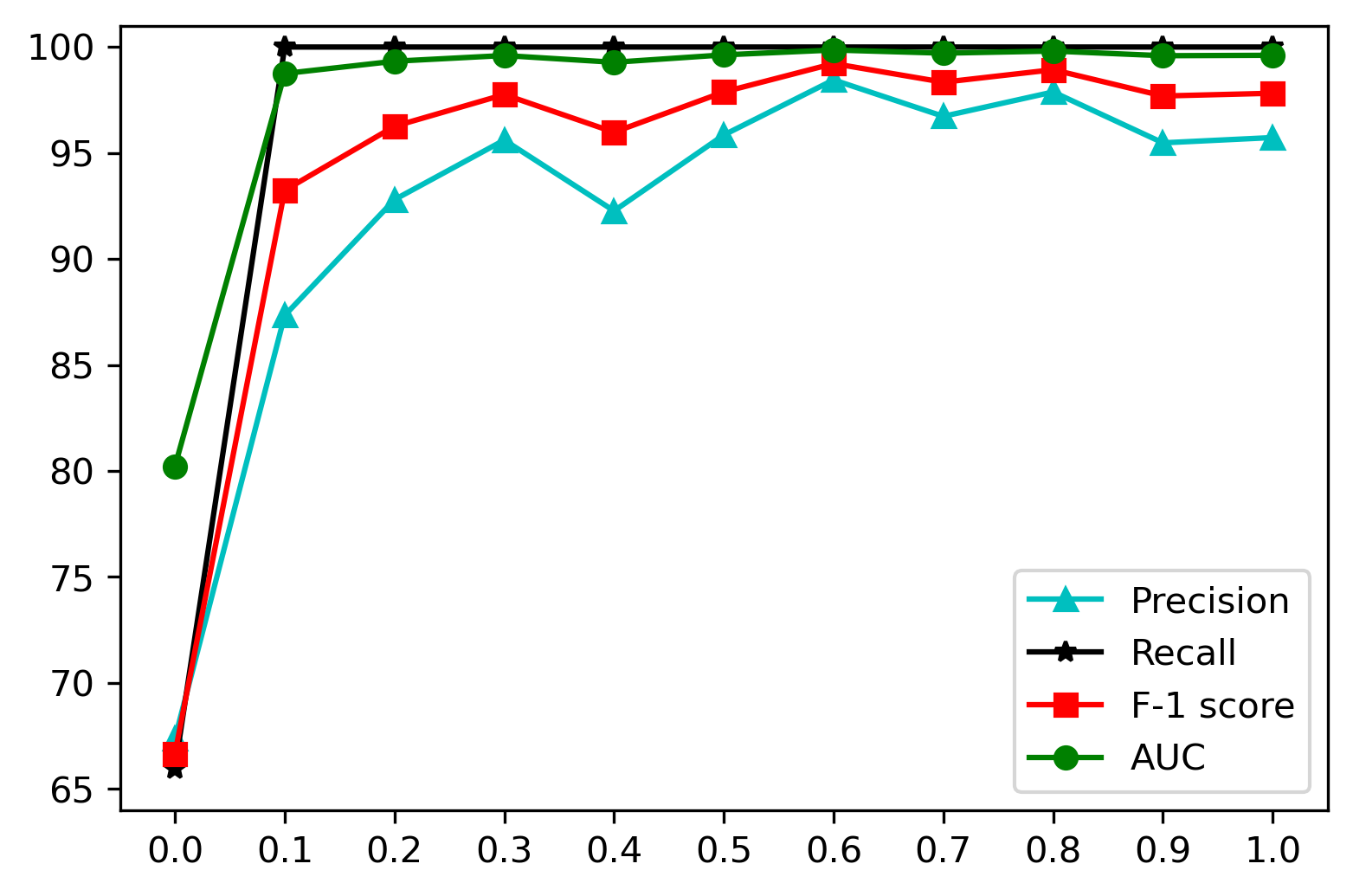}
       \caption{Triplet param. $\lambda_t$ in Eq. \ref{eq:triplet}}
       \label{fig:lambda_t}
   \end{subfigure}
    \hfill
    \begin{subfigure}[b]{0.23\textwidth}
       \centering
       \includegraphics[width=0.95\textwidth]{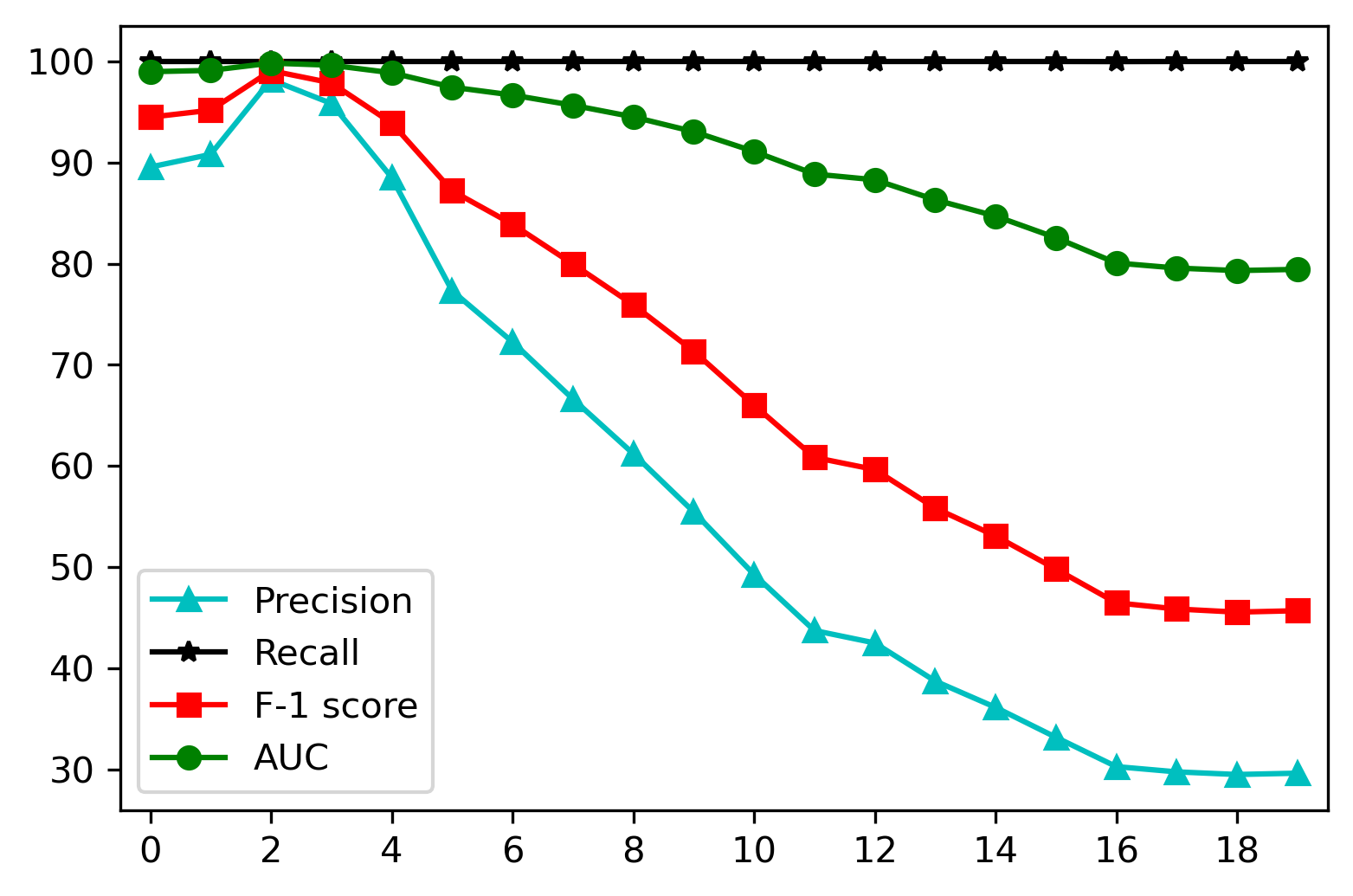}
       \caption{Cont. param. $\lambda_c$ in Eq. \ref{eq:continuity}}
       \label{fig:lambda_c}
   \end{subfigure}
    \hfill
    \begin{subfigure}[b]{0.23\textwidth}
       \centering
       \includegraphics[width=0.95\textwidth]{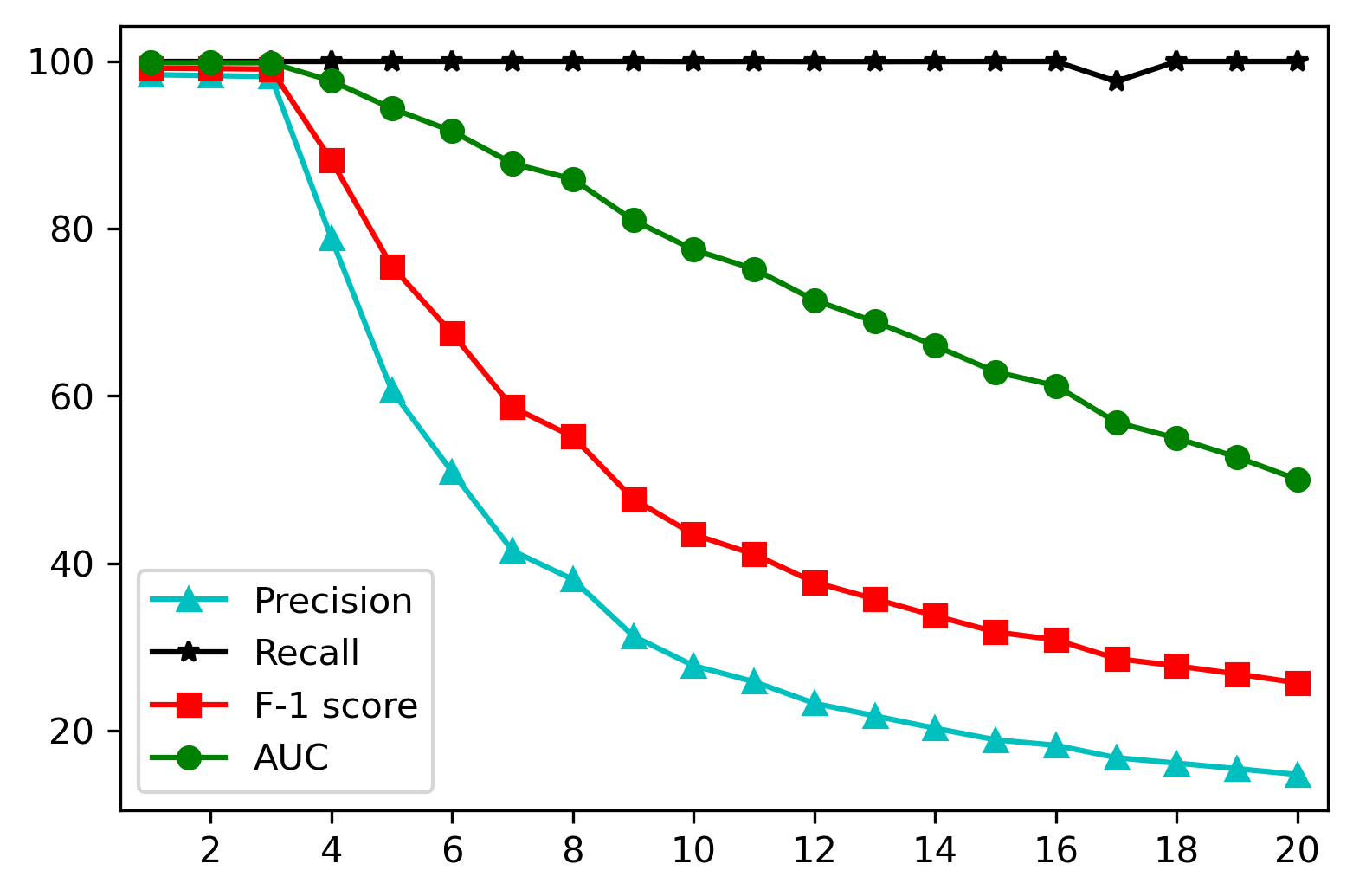}
       \caption{Sparsity param. $\lambda_s$ in Eq. \ref{eq:sparsity}}
       \label{fig:lambda_s}
   \end{subfigure}
   \caption{Hyperparameter sensitivity on the constant terms in losses defined in Equations \ref{eq:triplet} -- \ref{eq:sparsity}}
   \label{fig:sensitive-TB}
\end{figure}

{\bf \noindent Sensitivity analysis on the constant terms in loss terms defined in Equations \ref{eq:triplet} -- \ref{eq:sparsity}.}
We then analyze the sensitivity of the triplet parameter $\lambda_t$ (Equation \ref{eq:triplet}), continuity parameter $\lambda_c$  (Equation \ref{eq:continuity}) and sparsity parameter $\lambda_s$ (Equation \ref{eq:sparsity}). $\lambda_t$ indicates the margin between the distance from $\mathbf{r}_{z^+}$ to $\mathbf{c}$ and distance from $\mathbf{r}_{z^-}$ to $\mathbf{c}$. 
$\lambda_c$  indicates the expected number of the anomalous subsequences. For example, in our experiments, the sequence length is 20. $\lambda_c=0$ means all entries in the anomalous sequence are anomalous, while $\lambda_c=19$ means no consecutive anomalous entries in the sequence. $\lambda_s$  indicates the expected number of anomalous entries. $\lambda_s=1$ indicates we expect only one anomalous entry in the sequence. 

Figure \ref{fig:lambda_t} shows that when the margin is 0 ($\lambda_t=0$), CFDet cannot get a reasonable result, which meets the expectation that we need to ensure the detected anomalous subsequence have a large distance to the center. Once the margin is greater than 0 ($\lambda_t>0$), CFDet achieves much better performance. Figures \ref{fig:lambda_c} and \ref{fig:lambda_s} show the performance of anomalous entry detection with values ranging of $\lambda_c$ from 0 to 19 and $\lambda_s$ from 1 to 19, respectively.
We observe that when $\lambda_c$ and $\lambda_s$ are small, our approach achieves very good performance. On the other hand, with the increase of $\lambda_c$ and $\lambda_s$, the performance becomes worse. This is because for Thunderbird, the ratio of anomalous entries is small (0.15). Therefore, once we set $\lambda_c$ and $\lambda_s$ in a reasonable range, our approach can achieve good performance. On the other hand, if we keep increasing the values $\lambda_c$ and $\lambda_s$, the performance will get worse. This is because a large $\lambda_c$ or $\lambda_s$ would force the model to select more entries as anomalous, leading to high false alerts.

\begin{figure}[h!]
    \centering
  \begin{subfigure}[b]{0.23\textwidth}
        \centering
        \includegraphics[width=0.95\textwidth]{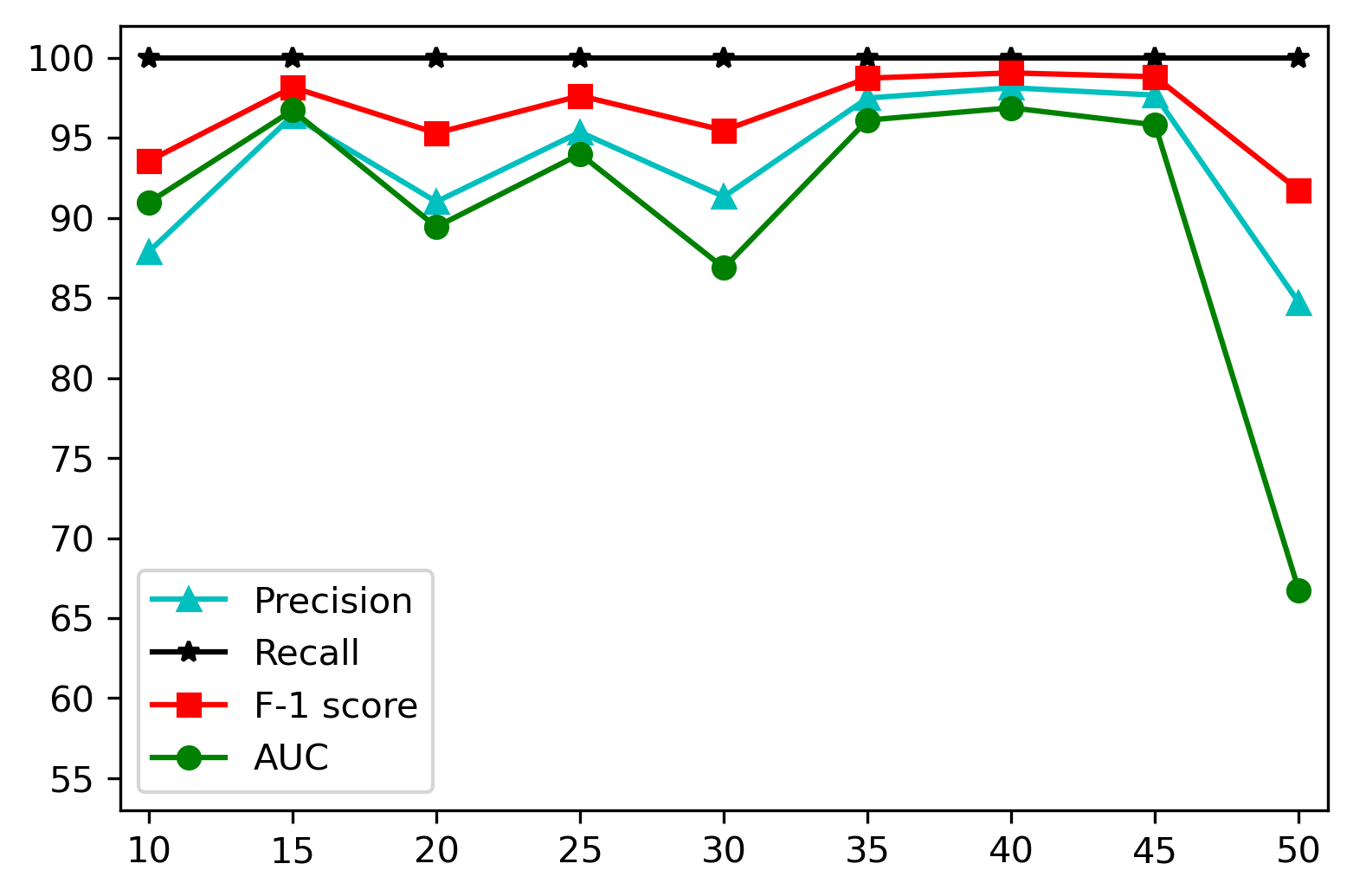}
        \caption{Anomalous Sequence Detection on $\mathcal{U}$}
        \label{fig:length_seq}
    \end{subfigure}
    \hfill
    \begin{subfigure}[b]{0.23\textwidth}
      \centering
        \includegraphics[width=0.95\textwidth]{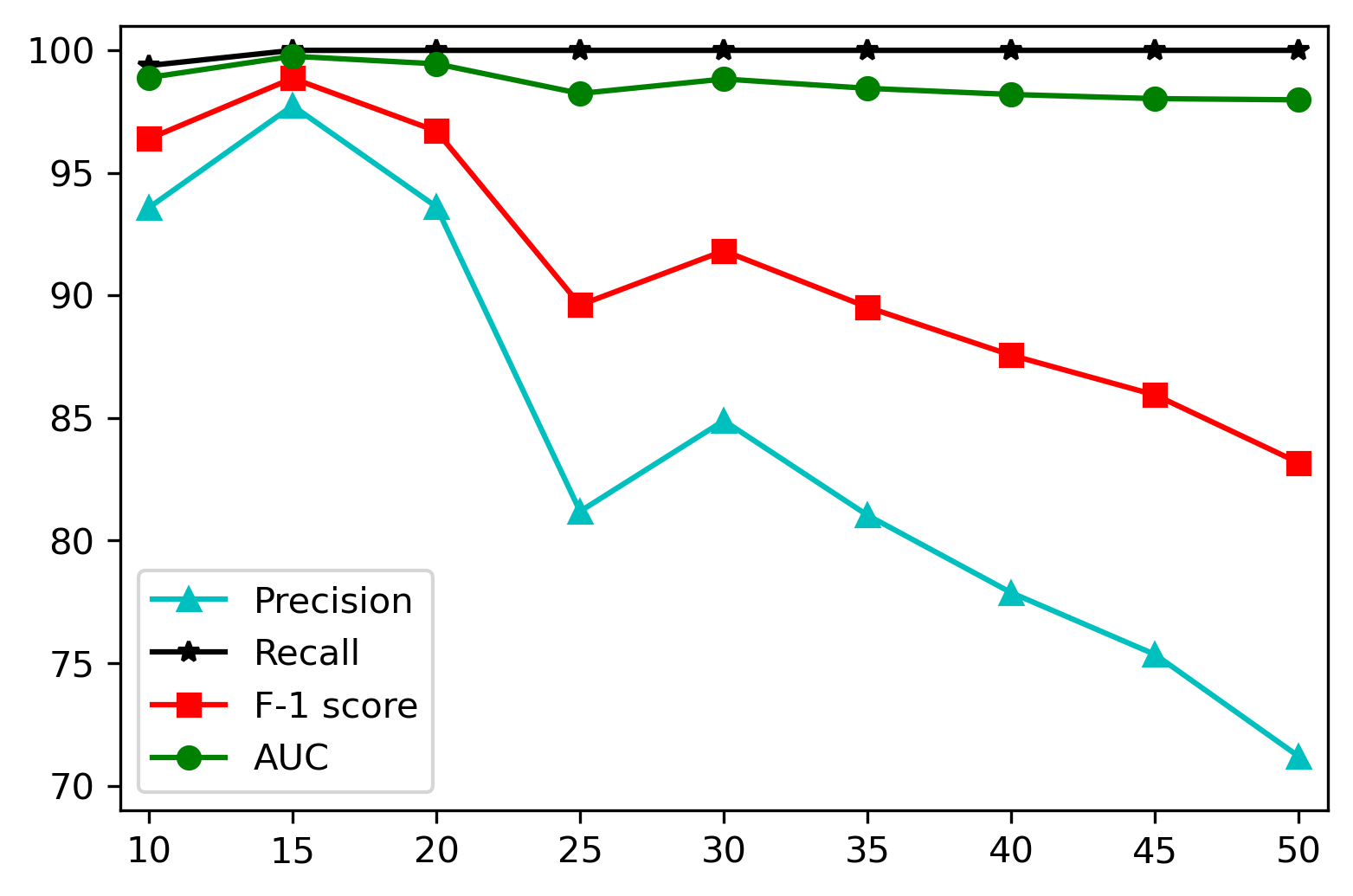}
      \caption{Anomalous Entry Detection on $\tilde{\mathcal{U}}^-$}
      \label{fig:length_entry}
  \end{subfigure}
  \caption{Performance of anomalous sequence and entry detection with various sequence lengths on  Thunderbird.}
  \label{fig:length}
\end{figure}

\begin{figure*}[ht]
    \centering
  \begin{subfigure}[b]{0.25\textwidth}
        \centering
        \includegraphics[width=\textwidth]{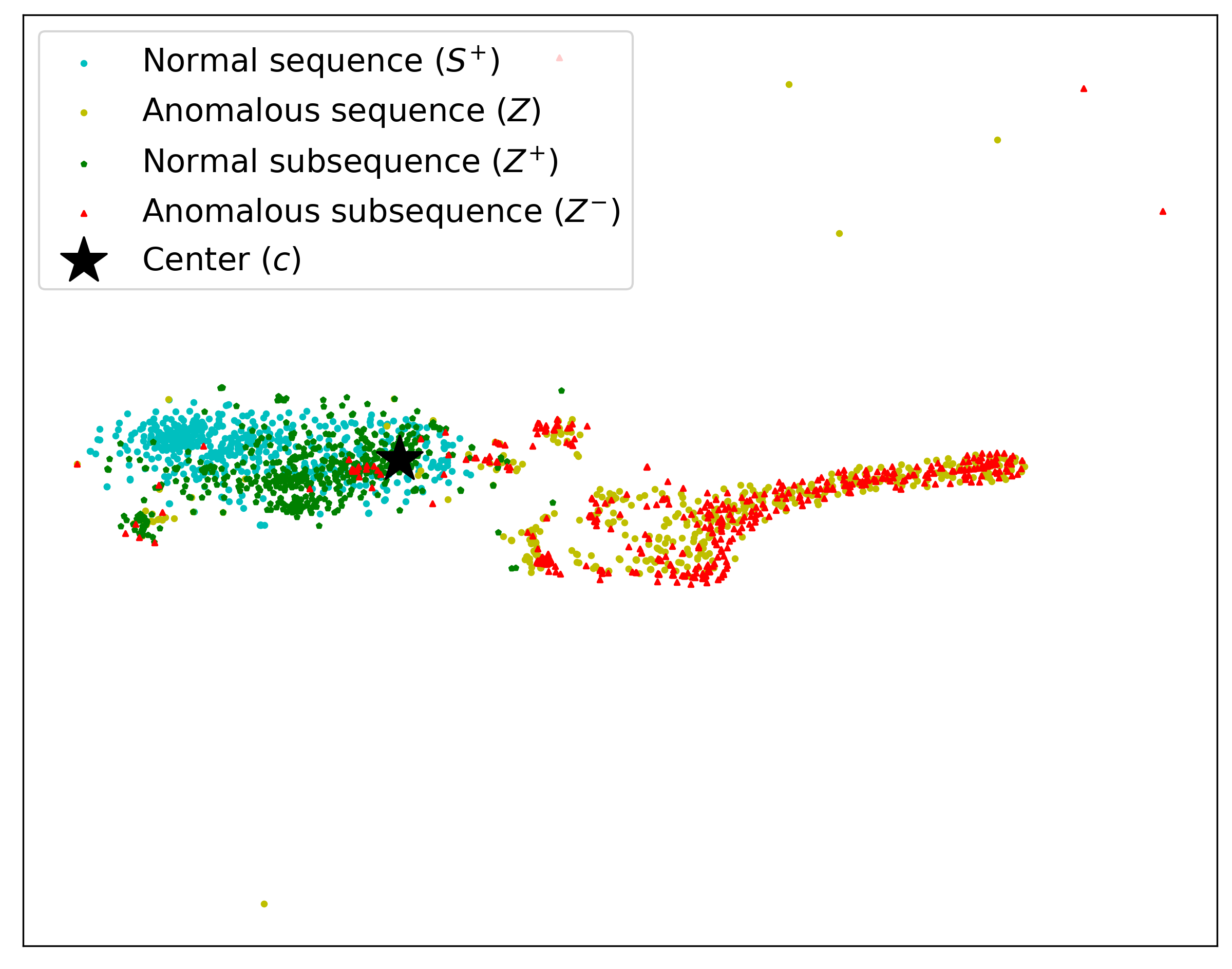}
        \caption{BGL}
        \label{fig:vis_bgl}
    \end{subfigure}
    \begin{subfigure}[b]{0.25\textwidth}
       \centering
       \includegraphics[width=\textwidth]{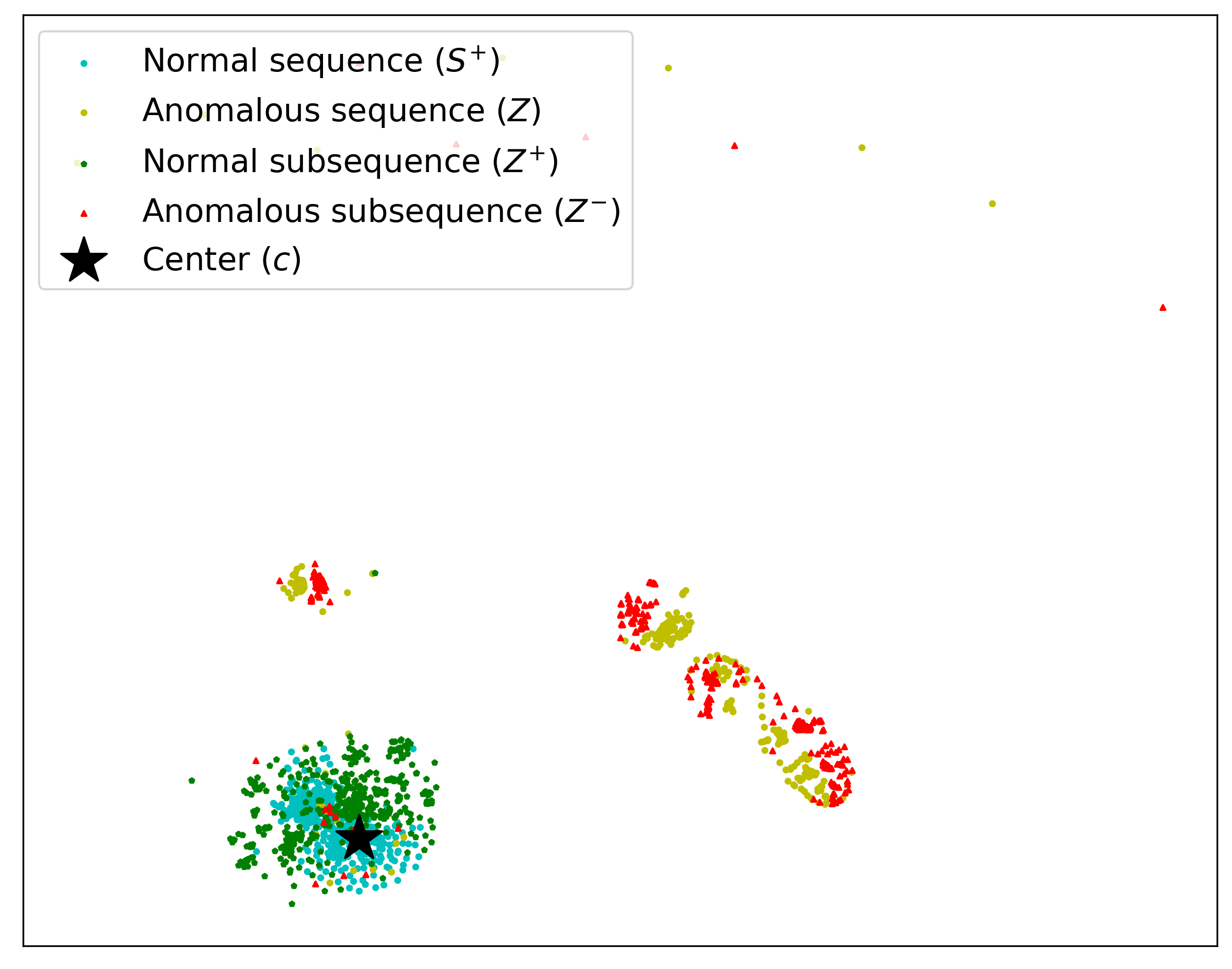}
       \caption{Thunderbird}
       \label{fig:vis_thunderbird}
   \end{subfigure}
    \begin{subfigure}[b]{0.25\textwidth}
      \centering
      \includegraphics[width=\textwidth]{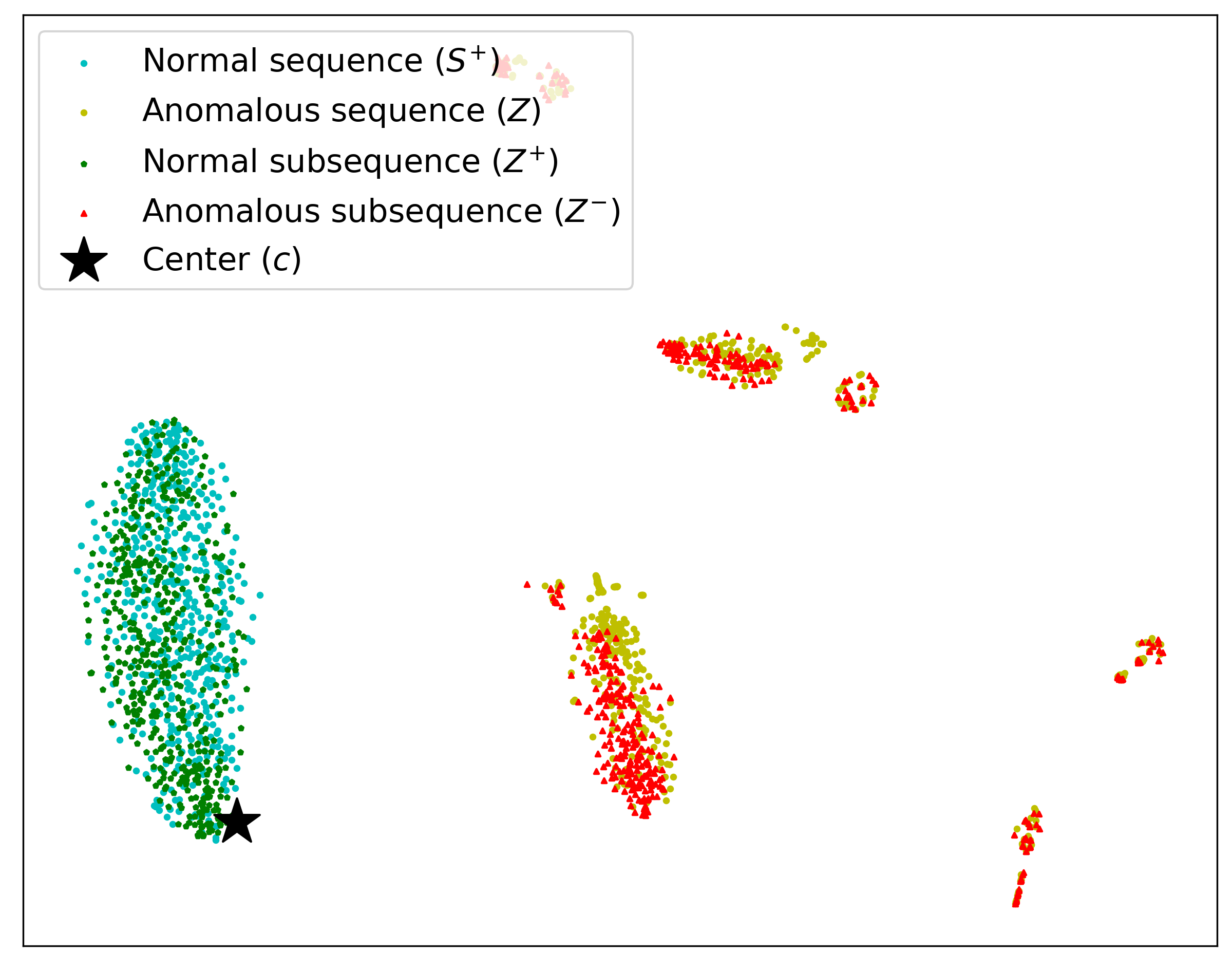}
      \caption{CERT}
      \label{fig:vis_cert}
    \end{subfigure}
     \hfill
   \caption{Visualization of the center (black), detected normal sequences $S^+$ (cyan), detected anomalous sequences $Z$ (yellow) and the corresponding normal $Z^+$ (green) and anomalous $Z^-$ (red) subsequences.}
   \label{fig:visualization}
\end{figure*}

\subsubsection{Various Sequence Lengths}
In our experiments, the default sequence length is 20. We further analyze the performance of our framework on the anomalous sequence and entry detection with various sequence lengths on the  Thunderbird  dataset. Figure \ref{fig:length} shows the evaluation results.

Figure \ref{fig:length_seq} shows the experimental results of anomalous sequence detection based on Deep SVDD. We can observe that the performance for the anomalous sequence detection keeps stable when the sequence lengths range from 10 to 45 for Thunderbird. When the sequence length is 50, the detection results from Deep SVDD got decreased. Figure \ref{fig:length_entry} shows the results of anomalous entry detection on the detected anomalous sequence set $\tilde{\mathcal{U}}^-$.  We can observe that the recall and AUC values are high when the sequence lengths increase from 10 to 50, but the precision as well as the F-1 scores keep reducing once the length is great than 15. It means on the Thunderbird dataset, CFDet is able to identify all the abnormal entries but predict some normal entries as abnormal (false positive) when the sequence length is large.

\subsubsection{Visualization}
We further visualize representations of normal and anomalous sequences in BGL, Thunderbird, and CERT datasets. We randomly select 500 detected normal and anomalous sequences, respectively. 
For the detected anomalous sequence $Z$ from each dataset, we then get the anomalous subsequence $Z^-$ and the normal subsequence $Z^+$. After deriving the representations of $S^+$, $Z$, $Z^-$, and $Z^+$ based on $f(\cdot)$, we adopt the t-SNE \cite{van2008visualizing} to map the representations into a two-dimensional space. Figure \ref{fig:visualization} shows the visualization plots. Overall, as expected, $S^+$ and $Z^+$ as normal sequences and subsequences are grouped together and located around the normal center $\mathbf{c}$, while $Z$ and $Z^-$ as anomalous sequences and subsequences are close but far from the normal center. It also demonstrates the detected $Z^+$ only contains the normal entries, while the corresponding $Z^-$ has the anomalous entries. Meanwhile, on both Thunderbird and CERT, the anomalous sequences and the corresponding subsequences are much more diverse compared with normal sequences.

\begin{figure}[h!]
    \centering
      \centering
      \includegraphics[width=0.48\textwidth]{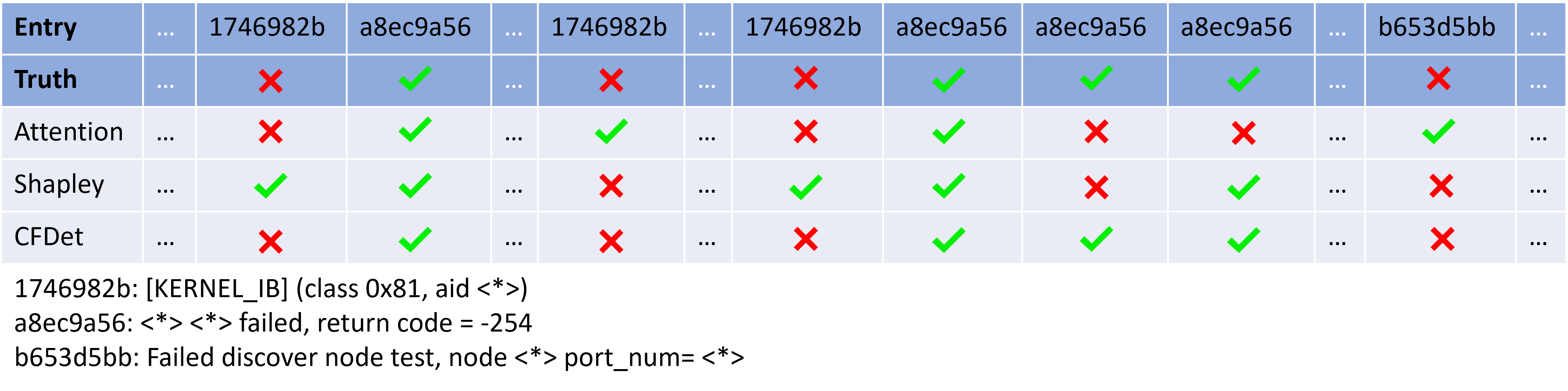}
  \caption{Case study of abnormal log entry detection on Thunderbird. Checkmark indicates an anomalous entry while X mark indicates a normal entry. }
  \label{fig:case-tb}
\end{figure}

\subsubsection{Case Study: Detecting Anomalous Log Entries}
We further conduct a case analysis on Thunderbird for anomalous log entry detection. In the Thunderbird dataset, CFDNet can detect all anomalous entries (100\% recall) with high precision. Hence, as shown in Figure \ref{fig:case-tb}, we can notice that CFDNet correctly labels all four log messages about fatal errors as anomalous entries and correctly predict other entries as normal. On the other hand, the two baselines, Attention and Shapley, cannot label all the anomalous entries and also mislabel some normal entries as anomalous.

\section{Conclusion}
In this paper, we have developed CFDet for fine-grained anomaly detection in sequential data based on the idea of counterfactual interpretation. CFDet is able to identify anomalous entries in sequences without using any labeled anomalous sequences/entries for training. In our CFDet, we first build an anomalous sequence detection model based on normal sequences to detect the anomalous sequences. We then develop a self-supervised anomalous entry detection model to identify the anomalous entries. The core idea is that the anomalous subsequence in the anomalous sequence should be far from the center of normal samples while the rest normal subsequence, considered as the counterfactual sequence of the original anomalous sequence, should be close to the center. Experiments on three datasets show that our model can identify anomalous entries with high accuracy. A potential future direction is to design an end-to-end anomalous entry detection approach from sequences without relying on the performance of anomalous sequence detection. For some types of sequential data, such as time series data, the data distributions are governed by the underlying causal structure. Another interesting direction is to consider the causal relationships when deriving the counterfactual explanations.


\end{document}